\documentclass[10pt,letterpaper]{article}
\usepackage[margin=1in, footskip=0.5in]{geometry}  
\usepackage{amsmath, amssymb} 
\usepackage[pdftex]{graphicx}
\usepackage{cite} 
\usepackage[right]{lineno}
\usepackage{changepage}
\usepackage{authblk} 

\usepackage{caption}
\usepackage[ruled,vlined]{algorithm2e}
\usepackage{booktabs}
\usepackage{tabularx}
\usepackage{array}
\usepackage{threeparttable}
\usepackage{float}

\def\im{inertial maneuvering}

\usepackage{mathtools}
\usepackage[table]{xcolor}
\usepackage{verbatim}

\title{\raggedright\textbf{Jointed tails enhance control of three-dimensional body rotation}}

\date{}

\author[1]{\raggedright Xun Fu}
\author[1]{\raggedright Bohao Zhang }
\author[2]{\raggedright Ceri J. Weber}
\author[2]{\raggedright Kimberly L. Cooper} 
\author[1,3]{\raggedright Ram Vasudevan}
\author[1,3,4]{\raggedright Talia Y. Moore\thanks{taliaym@umich.edu}}

\affil[1]{\raggedright\small Robotics, University of Michigan, Ann Arbor, MI, USA}
\affil[2]{\raggedright\small Department of Cell and Developmental Biology, University of California, San Diego, CA, USA}
\affil[3]{\raggedright\small Mechanical Engineering, University of Michigan, Ann Arbor, MI, USA}
\affil[4]{\raggedright\small Ecology and Evolutionary Biology, Museum of Zoology, University of Michigan, Ann Arbor, MI, USA}

\begin{document}
\raggedright
\setlength{\parindent}{0.5cm} 

\maketitle
\noindent \textbf{\large Abstract}\\
\vspace{5pt}
\noindent 
Tails used as inertial appendages induce body rotations of animals and robots---a phenomenon that is governed largely by the ratio of the body and tail moments of inertia.
However, vertebrate tails have more degrees of freedom (e.g., number of joints, rotational axes) than most current theoretical models and robotic tails.
To understand how morphology affects inertial appendage function, we developed an optimization-based approach that finds the maximally effective tail trajectory and measures error from a target trajectory.
For tails of equal total length and mass, increasing the number of equal-length joints increased the complexity of maximally effective tail motions.
When we optimized the relative lengths of tail bones while keeping the total tail length, mass, and number of joints the same, this optimization-based approach found that the lengths match the pattern found in the tail bones of mammals specialized for inertial maneuvering.
In both experiments, adding joints enhanced the performance of the inertial appendage, but with diminishing returns, largely due to the total control effort constraint.
This optimization-based simulation can compare the maximum performance of diverse inertial appendages that dynamically vary in moment of inertia in 3D space, predict inertial capabilities from skeletal data, and inform the design of robotic inertial appendages.
\bigskip

\section*{Introduction}
As exemplified by geckos and agamid lizards, tails can be swung to generate body rotations during aerial maneuvers to rapidly self right and transition from horizontal to vertical movement \cite{jusufi2008active,libby2012tail}.
Using physics-based models of this phenomenon, which we call in this paper \im\, we can even estimate the extent to which extinct or unsampled extant animals could have used inertial appendages to control body rotations \cite{libby2012tail}.
Drawing inspiration from these biological examples, roboticists have designed inertial appendages (a rigid rod moving in one plane of rotation) to facilitate body rotations in both aerial and terrestrial settings.
For instance, an active tail that controlled the aerial pitch rotations of a robot ensured safe landings after driving off a ramp \cite{chang2011lizard,libby2012tail}.
An insect-scale robot used an inertial tail to reorient during flight \cite{singh2019rapid}.
Similar studies demonstrated that a mobile robot with an active tail could turn at higher speeds compared to a tailless counterpart \cite{storms2016dynamic, pullin2012dynamic,casarez2013using}.
Other inertial robotic tails aid in torso stabilization \cite{briggs2012tails,zhao2015msu}.
These planar rigid tails have served as physical models to test biological hypotheses, enhance robot maneuverability, and are simple to design and control \cite{jusufi2010righting}.

However, vertebrate tails are far more diverse and complex than a simple rigid pendulum: they are made of a series of vertebrae that can bend in the dorsiventral and lateral planes with respect to each other \cite{laven2020measuring}.
Furthermore, increasing the number of rigid links likely affects the capability of a tail to form complex curvatures, which may facilitate dynamic changes in tail moment of inertia (MOI).
Thus, the maximum performance of a tail as an inertial appendage is likely to be greatly determined by its morphology.

Upon surveying various mammal tails, we found that tails with the same relative tail length could be comprised of different numbers of bones.
Having more bones within the same length of tail has been shown to enhance the inertial maneuvering effectiveness of a simulated tail robot swung from side to side in 2D \cite{rone2016dynamic}, but the simple trajectories used in this study may not correspond to the maximally effective trajectory for each morphology.
In light of the diverse tail behaviors and tail morphologies in the animal world, we first explored how to compare the maximum potential effectiveness of diverse inertial appendages.
We approached this question by developing physics-based simulations and applying optimization techniques to solve for the trajectory that maximizes the effectiveness of a given morphology. 

For such an optimization problem, 
among the most important constraints to consider are the control input limits.
Both in animals and in robotic systems, there are practical limits to the speed of actuation and sensory feedback as well as the complexity of control signals \cite{ijspeert2008central}.
If tails with different degrees of freedom (DOF) are subject to different total control input limits, we might erroneously find that tails with more links inherently perform better than tails with fewer links.
Therefore, a fair assessment requires that all tail configurations operate within similar control limits which ensures that any observed differences in performance are truly due to the design variations, not unequal control capabilities.

Here, we introduce an optimization-based approach to evaluate the maximum performance of inertial appendages under the same set of constraints.
This method finds the trajectory that minimizes the error (magnitude of distance, summed through the duration of the trial) to a target set of body rotations for each unique appendage configuration.
We found that, when comprised of equally sized, independently actuated links---with the same total tail length and mass---increasing the number of links increases performance of the inertial appendage.

In our survey of mammal tail morphology, we also found that the bone lengths vary among different regions within a tail.
This observation prompted us to wonder whether there was a functional benefit to this variability.
By allowing the relative size of links within a tail to vary, while maintaining the same total tail length and mass, we sought to determine whether regional changes in vertebral lengths further enhance \im.
We found that the optimal configuration consists of short links in the proximal and distal portions, with the longest links in the mid-region of the tail.
Surprisingly, the results of our variable link length optimization converged on the exaggerated crescendo and decrescendo pattern of vertebral lengths found in the tails of mammals associated with inertial maneuvering, compared to those without obvious specialized tail functions.
This concordance between the morphological data and the optimization results suggests that the exaggerated crescendo and decrescendo in vertebral lengths likely enhances the inertial maneuvering capability of a given tail morphology.

This optimization-based simulation is a generalized approach to estimate the function of any inertial appendage, regardless of shape.
This can be applied to estimate the potential \im~ability of vertebrate tails from simple series of skeletal measurements, compare inertial appendages that differ greatly in morphology and complexity, and provide essential guidance for the design of novel robotic inertial appendages.
 
\section*{Methods}
We isolated the effect of tail morphology on inertial maneuvering by simulating a 3D environment without gravity, wind resistance, or contact with a substrate.
Then, we developed models composed of two components: a torso and a tail.
In this context, the torso rotation can only be induced by movement of the tail.

We measured the maximal performance of a tail by using its actuation to make the torso follow a target ``trajectory'' of torso orientations.
The target torso trajectories involve large accelerations within a short time, resulting in pitch, roll, and yaw rotations similar to those animals experience when banking for a turn or preparing for a leap.
The task was formulated as a trajectory optimization problem.
By solving the problem, we evaluated the performance of tails with different configurations.
Note that the torso is not translating through space in the simulations, simply rotating.
Performance of the tail was then quantified by assessing how well the torso tracked the target trajectories through the actuation of the tail under certain constraints.
To quantify the tracking performance of a model, we used the integral of squared error between the target trajectory and the realized trajectory output by the optimization.

\subsection*{Model design and parameters}
\begin{figure}[!ht]
\begin{adjustwidth}{-0in}{0in}
    \centering
    \includegraphics[width = 0.6\textwidth]{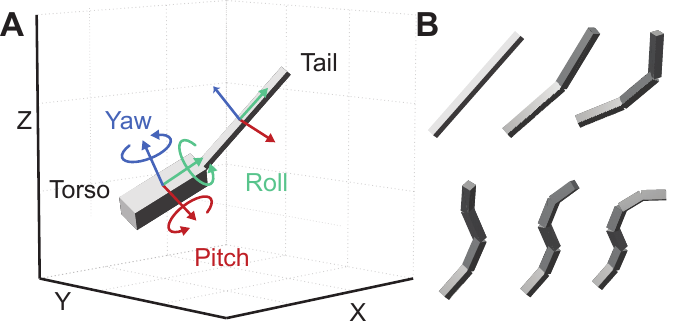}
    \caption{{\bf Model visualization.}
    An example of the model structure, featuring a torso and a single-vertebra tail, showing the axes of rotation for the torso and tail. 
    }
    \label{fig:robot_visual}
\end{adjustwidth}
\end{figure}

In our simulations, the torso and the tail were conceptualized as interconnected rigid bodies (see Fig \ref{fig:robot_visual}A), which facilitates visualization of rotations while simplifying and generalizing the morphology of diverse animals and robots.
The connection between these components was established through rotational joints.
Specifically, the torso was configured as a floating body represented by a uniformly dense square prism, virtually attached to the ground through an unactuated 3-DOF rotational joint, which enables roll, pitch, and yaw movements.

Throughout this study, we compared tails that all have equal total length and total mass, but differ in the number and size of bones that make up the tail.
The most simple tail, a single rigid pendulum, was represented by a uniformly dense square prism, and connected to the torso through a 2-DOF joint, which allow for pitch and yaw movements.
This represents a rigid tail with a single fused tail vertebra, or an animal that cannot create multiple inflexion points along the length of the tail.
In each jointed tail, the vertebrae (represented as rigid links) were represented as uniformly dense square prisms, each connected to each other and to the torso through the same type of 2-DOF joints.
Each DOF of these 2-DOF joints is actuated by an ideal, massless motor, supplying torques for active, independent pitch and yaw movements.
The exclusion of a direct roll DOF in the tail joints was based on three considerations: 
the relatively limited range of roll motion in the tails of real animals \cite{barbir2011effects,preuschoft2013torsion,orias2009rat}, 
the effectiveness of the combination of pitch and yaw movements in manipulating the torso's roll angle without a dedicated roll movement mechanism \cite{chu2023combining}; 
and the benefit of reducing the computational load by simplifying the model's joint complexity.

Detailed specifications of the models, including mass, size, joint torques, and other relevant parameters used for this study, are presented in Table \ref{tbl:rob_params}.
The ratios of torso mass to tail mass and torso length to tail length are similar to those observed in highly maneuverable animals, as discussed in \cite{libby2012tail}.

\begin{table}[!ht]
\begin{adjustwidth}{-0in}{0in}
\centering
\begin{threeparttable}
    \captionsetup{justification=raggedright, singlelinecheck=false, width=0.75\textwidth} 
    \caption{\bf Model parameters}
    \label{tbl:rob_params}
    \begin{tabularx}{0.75\textwidth}{X >{\centering\arraybackslash}p{0.3\textwidth}}
        \toprule
        \textbf{Property} & \textbf{Value} \\
        \midrule
        Torso Mass & 5 kg \\
        Tail Total Mass & 1.5 kg\\
        Torso Dimension (width×length×height) & 0.3 m × 1 m × 0.3 m \\
        Tail Dimension (width×length×height) &  0.1 m × 1.5 m × 0.1 m\\
        Tail Joint Velocity Range & $[-360, \, 360]$~deg/s\\
        Tail Joint Range of Motion & $[-60, \, 60]$~deg\\
        Tail Joint Torque Range & $[-5, \, 5]$ Nm\\
        \bottomrule
    \end{tabularx}
    \begin{tablenotes}
        \small
        \item The numbers in the table for dimensions follow the order of the local X, Y, and Z axes of the coordinate system illustrated in Fig. \ref{fig:robot_visual}, represented by red, green, and blue, respectively
    \end{tablenotes}
\end{threeparttable}
\end{adjustwidth}
\end{table}

We also assessed the influence of variations in vertebral lengths along the length of the tail.
In this case, the length of each tail vertebra can vary, so long as the total length of the tail remains the same.
Importantly, we leveraged the assumption that all vertebrae have a uniform density, so changes in vertebral lengths are accompanied by changes in mass, providing a more holistic understanding of the implications of size variations.
For models with non-uniform vertebral lengths in their tails, we modeled one to four vertebrae.
The results for the five-vertebra and six-vertebra configurations are not available because we have to employ symbolic representation to capture the dynamics of the models with non-uniform tail vertebra segment lengths.
As the system dimension grows, the process of exporting these symbolic representations for optimization use becomes considerably challenging.

\subsection*{Torso trajectory generation}
\begin{figure}[!ht]
    \centering
    \includegraphics[width = \textwidth]{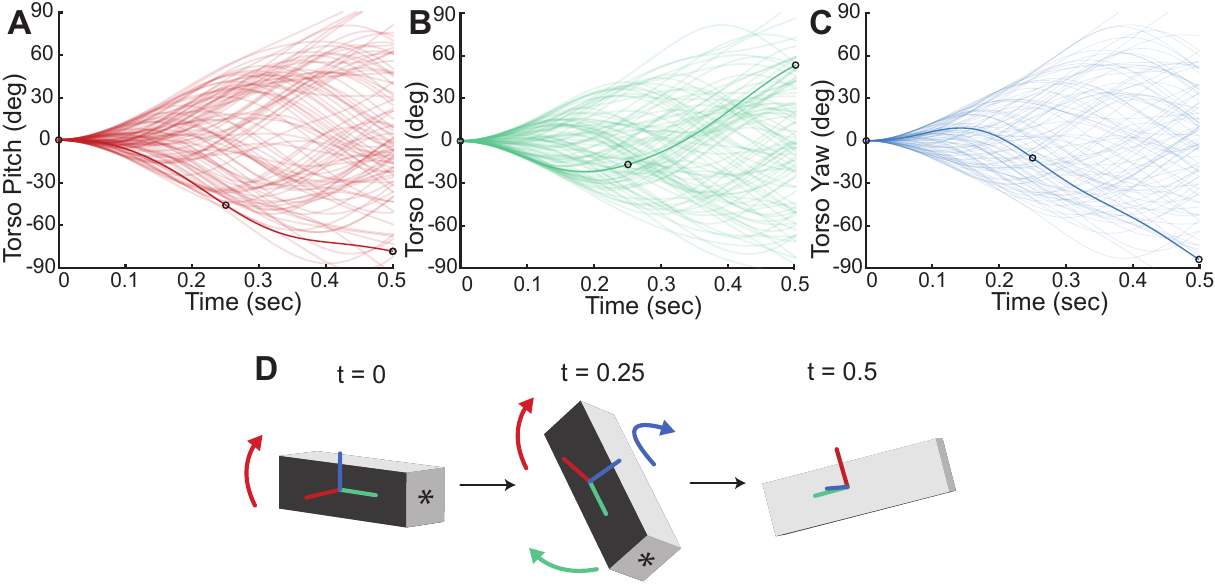}
    \caption{{\bf Goal torso rotation trajectories.}
    (\textbf{A, B, C}) depict the 100 target trajectories for torso pitch, roll, and yaw angles, respectively. 
    Each line represents an individual trajectory.
    Circles indicate the time points represented in (\textbf{D}).
    (\textbf{D}) depicts the torso orientation during three time points throughout one trajectory, bolded in (\textbf{A, B, C}).
    Curved colored arrows show the primary rotations that result in the differences between subsequent time steps.
    The asterisk indicates where the tail would attach.
    }
    \label{fig:torso_traj_range}
\end{figure}

We represented the torso's roll, pitch, and yaw rotational trajectories using the Fourier Series.
Such rotations are similar to those that animals would experience during banked turning \cite{jindrich2007mechanics}, self-righting \cite{jusufi2008active}, or perching \cite{boerma2019wings}.
A set of 100 trajectories for pitch, roll, and yaw (see Fig \ref{fig:torso_traj_range}), each with a duration of $0.5$ seconds were generated.
To accommodate the physical limitations, the roll, pitch, and yaw angles were constrained within the range of $[-180, 180]$~deg and angular velocities are limited to a range of $[-360, 360]$~deg/s for all 100 torso orientation trajectories.

Here, let $t \in [t_0, t_f]$ represent time.
We denote the torso’s orientation (i.e., roll, pitch, and yaw angles) by $\Theta: [t_0, t_f] \to \mathbb{R}^3$,
\begin{equation}
    \Theta(t) = \begin{bmatrix}\theta_1 (t), \; \theta_2 (t), \; \theta_3 (t)\end{bmatrix}^{\top} \label{eq:torso_ori_def}
\end{equation}
We generate random trajectories for the torso’s orientation using a fifth-degree Fourier series as follows:
\begin{equation}
    \theta_i (t) = a_{i0} + \sum_{j=1}^{5} (a_{ij} \cos(j\omega_i t) + b_{ij} \sin(j\omega_i t)) \label{eq:fs_def}
\end{equation}
where $\{a_{ij}\}_{j=0}^{5}$, $\{b_{ij}\}_{j=1}^{5}$, and $\omega_i$, for each $i = 1, 2, 3$, are adjustable parameters.
Different combinations of the $a_{ij}$ and $b_{ij}$ coefficients are sampled randomly to generate each of the 100 trajectories.
These coefficients are chosen to ensure that each Fourier series starts from zero, remains bounded, and that its derivative also remains bounded throughout the trajectory, which preserves the torso angle limit and the torso angular velocity limit.

\subsection*{Trajectory optimization}
\subsubsection*{Uniform vertebral lengths}
Given a tail that consists of $n_l$ vertebrae, we define the joint configuration trajectory as $q: [t_0, t_f] \to \mathbb{R}^{n_q}$, where $n_q = 2n_l + 3$.
For the models in our study, the joint configuration comprises the position of each revolute joint in the tail, with the first three elements corresponding to the torso orientation angles.
The joint velocity and joint acceleration trajectories are denoted by $\dot{q}: [t_0, t_f] \to \mathbb{R}^{n_q}$, $\ddot{q}: [t_0, t_f] \to \mathbb{R}^{n_q}$, respectively.

The dynamics of the given model are governed by the following equation: 
\begin{equation}
    \ddot{q}(t) =  - M(q(t))^{-1}\Big( - H(q(t), \dot{q}(t)) + \tau(t) \Big) \label{eq:general_dynamics}
\end{equation}
where $\tau(t) \in \mathbb{R}^{n_q}$ represents the joint control inputs, $M(q(t)) \in \mathbb{R}^{n_q \times n_q}$ is the mass-inertia matrix, and $H(q(t), \dot{q}(t)) \in \mathbb{R}^{n_q \times n_q}$ represents the combined effects of Coriolis forces and centrifugal forces all at time $t$.

Note that, by default, this dynamics equation assumes that all the joints, including the floating joints of the torso, can supply joint torques independently.
To emphasize that the task involves only actuating the joints in the tails to induce torso rotations, we forcibly apply zero torque to the floating joints of the torso:
\begin{equation}
    \ddot{q}(t) =  M(q(t))^{-1} \Big( - H(q(t), \dot{q}(t)) + \begin{bmatrix}{O}_{1\text{x}3}, & u(t)^{\top}\end{bmatrix}^{\top}\Big) \label{eq:dynamics_wzero_torsoinput}
\end{equation}
where $u: [t_0, t_f] \to \mathbb{R}^{n_q-3}$ represents the torque inputs for the tail joints. 

To express the dynamics of the model in a first-order form, as required by standard trajectory optimization, we define the system state trajectory and system dynamics as:
\begin{equation}
    x = \begin{bmatrix} q \\ \dot{q} \end{bmatrix} \label{eq:sys_state}
\end{equation}

\begin{equation}
    \dot{x} = f(x, u) = \begin{bmatrix} \dot{q} \\ \ddot{q} \end{bmatrix} \label{eq:sys_dynamics}
\end{equation}
where $x: [t_0, t_f] \to \mathbb{R}^{2n_q}$, where $\dot{x}: [t_0, t_f] \to \mathbb{R}^{2n_q}$

We define the objective function as follows,
\begin{equation}
        \underset{x,u, \dot{u}}{\text{min}} \int_{t_0}^{t_f} {\left\lVert \, Cx(t) - \Theta(t) \, \right\rVert}_2^2 \; dt \label{eq:obj_fun}
\end{equation}
where $C \in \mathbb{R}^{3\times 2n_q}$ is a selection matrix used to extract the torso orientations from the system states, and $\Theta$ represents the target torso orientation trajectory.
Thus, the cost function measures the integrated tracking error between the actual torso orientations and the target torso orientations.

The optimization is subject to other constraints, including the joint position and joint velocity constraints:
\begin{equation}
        S_{lb} \leq x(t) \leq S_{ub} \qquad \forall t \in [t_0, t_f] \label{eq:state_con}
\end{equation}
where the values of the bounds can be found in Table \ref{tbl:rob_params}, and the system control input constraints:
\begin{equation}
        U_{lb} \leq u(t) \leq U_{ub} \qquad \forall t \in [t_0, t_f] \label{eq:input_con}
\end{equation}

In addition, to avoid generating solutions with impractical, erratic control inputs, we introduced constraints on the rates of change of the inputs:
\begin{equation}
         R_{lb} \leq \dot{u}(t) \leq R_{ub} \qquad \forall t \in [t_0, t_f] \label{eq:input_rate_con}
\end{equation}

Importantly, we also included a constraint that limits the control efforts of the tail joints:
\begin{equation}
         u(t)^{\top}u(t) \leq E  \qquad \forall t \in [t_0, t_f] \label{eq:total_control_effort_con}
\end{equation}
This constraint was applied to all the models, and the value of the bound $E$ was chosen such that it does not exceed the maximum possible value for the model with a single-vertebra tail.
This ensures that the models with different numbers of tail vertebrae have the same level of total control effort limit, facilitating an equitable comparison.

Lastly, we included constraints to prevent self-collisions:
\begin{equation}
          {g}(x(t)) \leq O_{n_g \times 1}  \qquad \forall t \in [t_0, t_f] \label{eq:no_self_coll_con}
\end{equation}
To prevent the tail from intersecting with the torso, we over-approximated the torso using spheres.
Similarly, the joint connections in the tail, as well as the tip of the tail, were approximated by spheres.
To avoid contact between the tail and the torso, we formulated constraints such that the distance between each sphere on the tail and each sphere on the torso was greater than the sum of their respective radii.
Each element in the function $g: [t_0, t_f] \to \mathbb{R}^{n_g}$ specifies one such relationship between a pair of spheres.
The dimension of this constraint, $n_g$, varies with the number of tail vertebrae.
It is important to note that, given the range of motion values and the lengths of the tail vertebrae, there is no need to explicitly consider self-collision among the tail vertebrae.
This simplification reduces the number of nonlinear constraints in the optimization problem, thereby reducing the computational load required to solve the problem.

Combining everything together, the complete formulation of the trajectory optimization problem is presented as follows:

\begin{align}
        & \underset{x, u}{\text{min}}
        && \int_{t_0}^{t_f} {\left\lVert \, Cx(t) - \Theta(t) \, \right\rVert}_2^2 \; dt \nonumber \\
        & \text{s.t.}
        && \dot{x}(t) = f(x(t), u(t)) && \forall t \in [t_0, t_f] \nonumber \\
        &&& S_{lb} \leq x(t) \leq S_{ub} && \forall t \in [t_0, t_f] \nonumber \\
        &&& U_{lb} \leq u(t) \leq U_{ub} && \forall t \in [t_0, t_f] \nonumber \\
        &&& R_{lb} \leq \dot{u}(t) \leq R_{ub} && \forall t \in [t_0, t_f] \nonumber \\
        &&& u(t)^{\top}u(t) \leq E && \forall t \in [t_0, t_f] \nonumber \\
        &&& g(x(t)) \leq O_{n_g \times 1} && \forall t \in [t_0, t_f] \nonumber \\
        \label{opt:traj_opt_uniform_len}
\end{align} 

Methods for solving trajectory optimization problems can be divided into two categories: indirect and direct \cite{betts2010practical}.
Here, we used the direct collocation method, which discretizes the continuous-time trajectory optimization problem by approximating the continuous functions in the problem statement as polynomial splines, thereby converting it into a nonlinear problem.
To provide a solution that achieves high-order accuracy without necessitating super fine intervals or a small time step, which typically increases the computational load, we employed the Hermite-Simpson collocation method. 
This high-order method approximates continuous functions as piece-wise quadratic functions, offering more accurate approximations. 

For conciseness, we omit the nonlinear program formulation transcribed from the continuous-time trajectory optimization.
Detailed tutorials on such transcription process can be found in \cite{betts2010practical, kelly2017introduction}.

\subsubsection*{Variable vertebral lengths}
To formulate a trajectory optimization problem where the tail vertebral lengths are variable, several modifications were made from the original trajectory optimization problem (\ref{opt:traj_opt_uniform_len}).

First the vertebral lengths, $L = \{l_i \in \mathbb{R}\}_{i=1}^{n_l}$, were included as decision variables.
Just as the system dynamics become a function of both the system states and the vertebral lengths, so does the function $g$ for preventing self-collisions.
Note that, the mass, moment of inertia of each vertebra was correlated to the vertebral length.
Meaning, any change made to the vertebral length resulted in the change in the mass and moment of inertia.

To ensure the optimization results from variable vertebral length tails would be comparable to the results from uniform vertebral length tails, we introduced a constraint to maintain the sum of the vertebral lengths $\Gamma$ (i.e., total tail length) constant. 
Furthermore, each vertebral length was assigned a positive lower bound to preclude negative values.
Here, we set this positive lower bound to $H_{lb} = 0.2$~m.
This choice is made to avoid the necessity of adding extra constraints for self-collision avoidance, which would be required to account for potential self-collision between the tail vertebrae with variable lengths, and scenarios where spheres over-approximating the joint connections might not intersect with the torso, yet the portion of the vertebra between the spheres could collide with the torso.
While it is possible to employ more comprehensive self-collision constraints, such as convex polytopes \cite{lien2007approximate}, for more nuanced collision avoidance, the substantial increase in implementation complexity and the potential increase in computational time render them unnecessary for this study.
It is also noteworthy that a 0.2~m lower bound is smaller than the uniform vertebral length of a model with a six-vertebra tail, thus allowing the exploration of shorter vertebrae not considered in the original optimization problem.

With these changes, the formulation of the trajectory optimization allowing varied tail vertebral lengths is as follows,

\begin{align}
        & \underset{x, u, L}{\text{min}}
        && \int_{t_0}^{t_f} {\left\lVert \, Cx(t) - \Theta(t) \, \right\rVert}_2^2 \; dt \nonumber \\
        & \text{s.t.}
        && \dot{x}(t) = f(x(t), u(t), L) && \forall t \in [t_0, t_f] \nonumber \\
        &&& S_{lb} \leq x(t) \leq S_{ub} && \forall t \in [t_0, t_f] \nonumber \\
        &&& U_{lb} \leq u(t) \leq U_{ub} && \forall t \in [t_0, t_f] \nonumber \\
        &&& R_{lb} \leq \dot{u}(t) \leq R_{ub} && \forall t \in [t_0, t_f] \nonumber \\
        &&& u(t)^{\top}u(t) \leq E && \forall t \in [t_0, t_f] \nonumber \\
        &&& g(x(t), L) \leq O_{n_g \times 1} && \forall t \in [t_0, t_f] \nonumber \\
        &&& l_i \geq H_{lb} \nonumber && \forall i \in {1, \dots, n_l} \\
        &&& \sum_{i=1}^{n_l} l_i = \Gamma \nonumber \\
        \label{opt:traj_opt_varied_len}
\end{align}
We also transformed this trajectory optimization problem into a nonlinear program.
For brevity, we omit the specifics of the nonlinear program here.

\subsection*{Optimization implementation}

We employed MATLAB's FMINCON as the solver for the nonlinear programs in the previous subsection.
To enhance computational efficiency, the calculation of model dynamics, which constitutes a significant portion of computation time, was performed using the Pinocchio C++ library\cite{carpentier2019pinocchio} and Roy Featherstone's algorithm \cite{featherstone2014rigid}, recognized for their efficiency in calculating rigid-body dynamics.
To improve solution optimality, analytical derivatives were provided for all involved constraints.
Regarding the discretization of continuous-time trajectory optimization over 0.5~s, a time step $dt$ of 0.004~s was selected, balancing precision with computational feasibility.

Given the nonlinear nature of the programs, well-chosen initialization can significantly improve the solver's effectiveness.
Conversely, a poor initial guess may lead to the solver getting trapped in a suboptimal local minimum.
To navigate this, we introduced five distinct initial conditions for each program.
For programs where the length of each vertebra segment in the tail of model was uniform, the initial conditions included: zero states and zero control inputs, simplistic inputs that nevertheless satisfy all constraints (including the dynamics constraints); a straight line in state space between the initial and final states for the torso orientation trajectory, a commonly used initialization strategy; and three other randomly generated initial conditions within the system state and control input boundaries.
For programs optimizing over tail vertebra segment lengths, we applied the same five distinct initial conditions for the states and control inputs.
For the length decision variables, we used uniform vertebral lengths as the initial guess.

Upon applying these various initial conditions, we assessed the results based on the cost associated with the objective function, selecting the ``best" solution characterized by the minimum objective function value.
We validated the solution for each program by forward simulating the model dynamics in MATLAB using ODE45, applying the control inputs derived from the optimization problem, and confirmed that the simulated states match the states derived from the optimization solutions, which provides a solid basis for the findings and conclusions drawn in subsequent subsections. 

\subsection*{Vertebral measurements}
We examined 19 mammal species for this study (see supplementary data).
For each species, we selected the most complete adult skeletal specimen available.
Age was determined by long bone and sacral epiphyseal fusion.
The tail begins with the first free vertebra immediately distal to the sacrum.
We used digital calipers to measure the distance between the most proximal and the most distal points of each cauldal vertebra centrum, near the mid-sagittal point of the element.
As the most complete tail skeletons were often articulated, these centrum lengths include the vertebral epiphyses.
To account for large differences in body size among species in the dataset, we normalized all vertebral lengths by the length of the first caudal vertebra, which we found correlated strongly with body size.

We examined two functional groups of mammals: those that are similar to generalist mice and those that are likely to use their tail as an inertial appendage.
Mouse-like species were selected from mammal families Muridae, Cricetidae, and Heteromyidae and Soricidae (shrews).
All species in this group are quadrupedal terrestrial mammals with elongate tails that have no obvious specialized function.
To determine whether a species use their tail as an inertial appendage, we either found relevant peer-reviewed literature or video clips on Youtube (see supplementary data).
Importantly, mammals in both groups have approximately the same body to tail moment of inertia ratio, indicating that aside from skeletal morphology, they should both be equally capable of the same degree of inertial maneuvering.
For species without a specific peer-reviewed paper explicitly studying tail movement, inertial maneuvering was deemed likely if an individual exhibited active and rapid tail motion during extended aerial phases (e.g., jumping from tree to tree).
To isolate inertial maneuvering from potential aerodynamic effects (as described in \cite{norby2021enabling}), species with fluffy tails (i.e., the silhouette of the tail bones were significantly altered by fur for more than 50\% of the length) were excluded from the analysis.
\section*{Results}
\subsection*{Uniform vertebral lengths}

\begin{figure}[!ht]
    \centering    \includegraphics[width =0.7\textwidth]{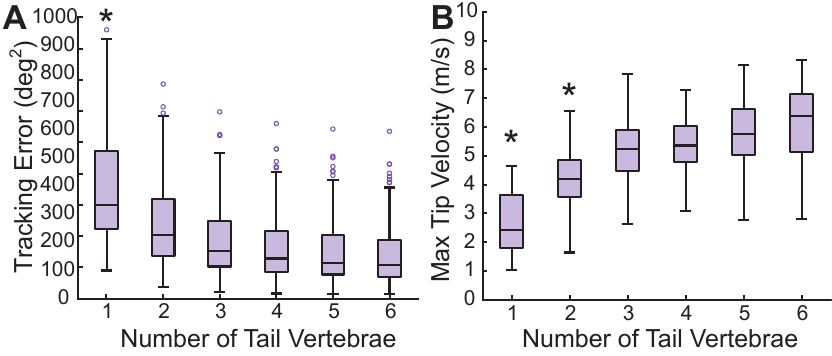}
    \caption{{\bf Each additional vertebra enhances inertial appendage performance.}
    (\textbf{A}) The tracking error decreases with each additional vertebra.
    The single-vertebra tail had significantly higher tracking error compared to all other tail morphologies (see Tab. \ref{tab:anova} A for significance levels).
     (\textbf{B}) The velocity of the tail tip increases with each additional vertebra, which increases the total torque generated by the tail.
    These magnitudes are computed by taking the norm of the 3D vectors representing the linear velocities at the tail tip (see Tab. \ref{tab:anova} B for significance levels).
    }
\label{fig:uniform}
\end{figure}

Motion in both pitch and yaw axes was exhibited by all tails, regardless of the number of vertebrae.
This tail motion is logical, considering that the target trajectories included rotations in all three axes.
With two vertebrae, the optimal trajectory included angular changes between the two vertebrae.
This indicates that, although it is a possible solution, coupling the two vertebrae together (equivalent to a single rigid link tail) is not as effective as creating a bend along the length of the tail.
This is likely due to the additional velocity experienced by the mass of the second vertebra, which thereby generates more torque on the tail-body joint.
For tails with three or more vertebrae, the optimal trajectories always included decoupled actuation of neighboring vertebrae, which resulted in complex curves with multiple inflexion points.
The velocity of the tip of the tail also increased with each additional vertebra (Fig. \ref{fig:uniform} B).

The average tracking error decreased as the number of vertebrae increased (Fig. \ref{fig:uniform} A).
Compared to the single-vertebra tail case, the six-vertebrae tail configuration improved tracking accuracy by 62.03\%.
This enhanced tracking accuracy demonstrates that an increase in the number of vertebrae improved the model's ability to trace a maneuverable trajectory, thereby suggesting that adding vertebrae to a tail increases the rotational maneuverability of the torso.
There was a significant effect of vertebral number on tracking error, according to a one-way ANOVA ($F_{5,93}=40.42,p=2*10^{-16},\alpha=0.05$, see Tab. \ref{tab:anova} for significance levels).

\begin{table}[h]
    \centering
     \caption{\textbf{Vertebra number greatly affects performance.}
     \textbf{A} The significance values for the effect of vertebral number on tracking error (as shown in Fig. \ref{fig:uniform} A).
     \textbf{B} The significance values for the effect of vertebral number on maximum tip velocity (as shown in Fig. \ref{fig:uniform} B).
     The pairwise significance values are computed by a Tukey Honest Significant Difference test.
     The numbers represent the number of uniform length vertebrae in the tail.
     The significance levels are depicted with symbols: `()' indicates $p<0.1$, `.' indicates $0.1>p>0.05$, `*' indicates $0.05>p>0.01$, `**' indicates $0.01<p<0.001$, `***' indicates $0.001<p<0$, `--' indicates a redundant comparison.}
    \begin{tabular}{c|cccccc}
            \textbf{A}&  1     & 2        & 3     & 4     & 5     & 6 \\
            \hline
        1   &  --    &  --      &   --  &  --   &  --   &  -- \\
        2   & $***$  & --       &   --  &  --   &  --   &  -- \\
        3   & $***$  &  ()       &--     &  --   &  --   &  -- \\
        4   & $***$  & $**$    & ()    &--     &  --   &  -- \\
        5   & $***$  & $***$    &  ()   & ()    &  --   &  -- \\
        6   & $***$  & $***$    & ()     & ()    &  ()   &  -- \\
    \end{tabular}
    \quad
    \begin{tabular}{c|cccccc}
            \textbf{B}&  1     & 2        & 3     & 4     & 5     & 6 \\
            \hline
        1   &  --    &  --      &   --  &  --   &  --   &  -- \\
        2   & $***$  & --       &   --  &  --   &  --   &  -- \\
        3   & $***$  &$***$     &--     &  --   &  --   &  -- \\
        4   & $***$  & $***$    & ()    &--     &  --   &  -- \\
        5   & $***$  & $***$    &$*$   & .    &  --   &  -- \\
        6   & $***$  & $***$    &$***$  & $***$  &  ()   &  -- \\
    \end{tabular}
    
    \label{tab:anova}
\end{table}

\begin{figure}[!ht]
    \centering
    \includegraphics[width=\textwidth]{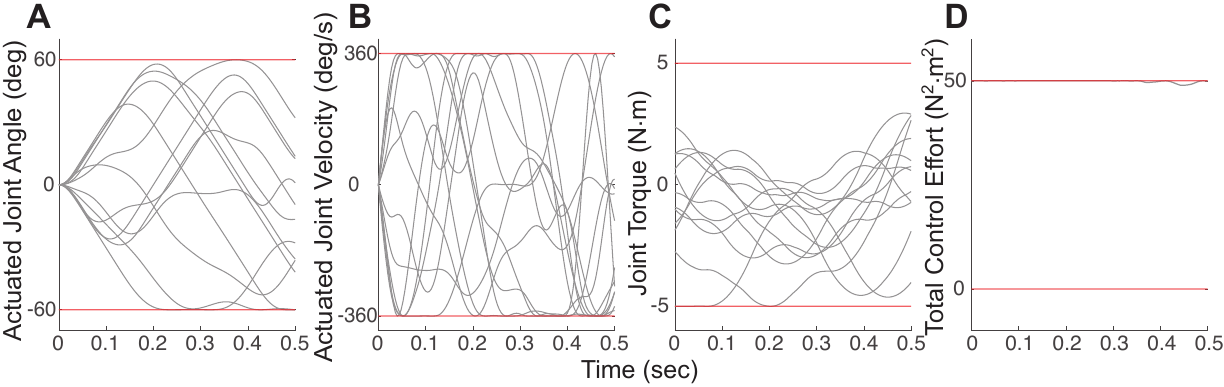}
    \caption{\textbf{Total control effort constraint limits performance of complex tails.}
    Each gray line represents the value of a physical parameter for one joint degree of freedom.
    The red lines represent the limit imposed for that parameter.
    (\textbf{A}) represents the actuated joint angle, 
    (\textbf{B}) represents the actuated joint velocity, 
    (\textbf{C}) represents the joint torque, and
    (\textbf{D}) represents the total control effort.
    Note that only in \textbf{D}, the gray lines are near the limit throughout the duration of the trial, indicating that total control effort is the primary constraint on performance in this trial.
    }
    \label{fig:constraints}
\end{figure}

While the tracking error consistently reduced with the increase in the number of vertebrae, the magnitude of improvement diminished.
To understand the source of this diminishing benefit, we examined the values of the physical constraints in each trial.
For a 6-link tail tracking a single trajectory (Fig. \ref{fig:constraints} A, B, C) actuated joint angles, velocities, and torques, the values for each joint DOF varied across the entire constrained range.
On the other hand, for this trial, the total control effort was near the limit for the entire duration of the trial (Fig. \ref{fig:constraints}).
An important difference to note between these parameters is that the total control effort constraint was based on the sum of control effort for all DOF in the model, whereas the other physical parameter constraints are imposed for each DOF independently.
Thus, because the maximum potential total control effort increased with each additional vertebra, the total control effort constraint is likely to be a more important limiting factor as the number of vertebrae increases, thereby causing the diminishing benefits of highly jointed tails in these simulations.

\subsection*{Variable vertebral lengths}

Tails with variable segment lengths also resulted in 3D tail motions that leveraged a combination of non-zero pitch and yaw angles across the rotational DOFs to reorient torso orientation.
In comparison to the uniform length vertebrae, tails with optimized vertebral lengths had higher local curvatures because two smaller vertebrae are nearly equivalent to one vertebra with twice the range of motion.
This level of complexity in curvature cannot be achieved with uniformly sized tail segments, highlighting the advantages of variable vertebra lengths in tail design for control of torso orientation.

\begin{figure}[H]
\begin{adjustwidth}{-0in}{0in}
    \centering
    \includegraphics[width = \textwidth]{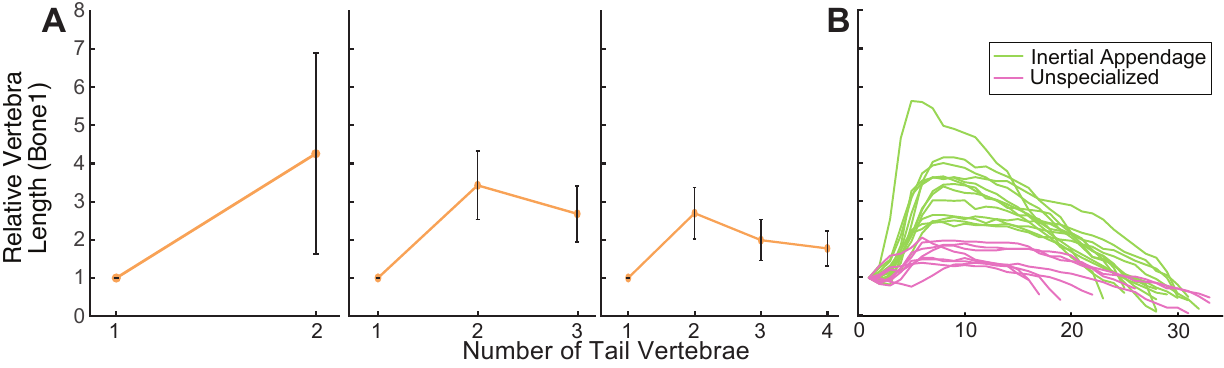}
    \caption{{\bf Optimized vertebral lengths resemble mammal tails specialized for inertial maneuvering.}
    (\textbf{A}) Optimized vertebral lengths for models with different tail configurations  demonstrate a crescendo and decrescendo pattern.
    (\textbf{B}) In comparison to mammals with unspecialized mouse-like tails, mammals that use their tails for inertial maneuvering have a more pronounced crescendo and decrescendo pattern in their tail vertebral lengths.}
    \label{fig:len_distribution}
\end{adjustwidth}
\end{figure}

The optimization on the lengths of individual vertebrae (within a tail of constant total length and mass) consistently resulted in differences in vertebral length along the length of the tail
(Fig \ref{fig:len_distribution} A).
In all tails, the first vertebra was always the shortest.
For configurations involving three and four vertebrae, the second vertebra was consistently the longest.
The third and fourth vertebrae decreased in size, but never reached the short length of the first vertebra.

\begin{figure}[!ht]
    \centering
    \includegraphics[width = 0.7\textwidth]{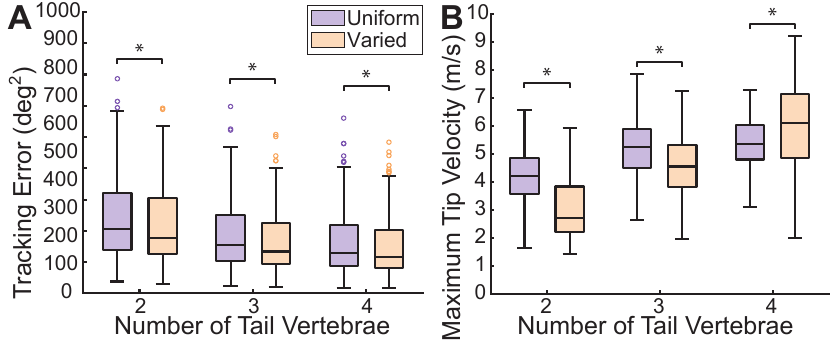}
    \caption{{\bf Comparison between uniform and variable vertebral lengths.}
    A) Tails with optimized vertebral lengths, rather than uniform vertebral lengths, show significant improvement in tracking error, regardless of the number of vertebrae in the tail.
    B) The magnitude of the maximum velocity of the most distal point on the tail increases as the number of vertebrae within a tail increases.
    The `*' symbol indicates significant pairwise differences at an $\alpha=0.05$ significance level, according to a paired t-test.
    }
    \label{fig:box_plot_track_err_varied_len}
\end{figure}

Tails with variable vertebral lengths exhibited lower tracking errors in comparison to tails with uniform vertebral lengths but the same number of total vertebrae, mass, and total tail length, (Fig \ref{fig:box_plot_track_err_varied_len} A).
This represents, on average, a 10.01\% improvement in tracking accuracy.
Allowing for variation in vertebral lengths within a tail resulted in a significant decrease in tracking error compared to the uniform vertebral length with the same number of vertebrae, regardless of the number of vertebrae (Paired t-test, 
$p=2.2*10^{-15}$, 
$2.2*10^{-16}$, 
$2.2*10^{-16}$; 
$t=9.4$,
$11.5$,
$11.2$, 
$DF=99$, for 2, 3, 4 vertebra tails, respectively, Fig. \ref{fig:box_plot_track_err_varied_len}).

\begin{figure}[!ht]
    \centering
    \includegraphics[width = \textwidth]{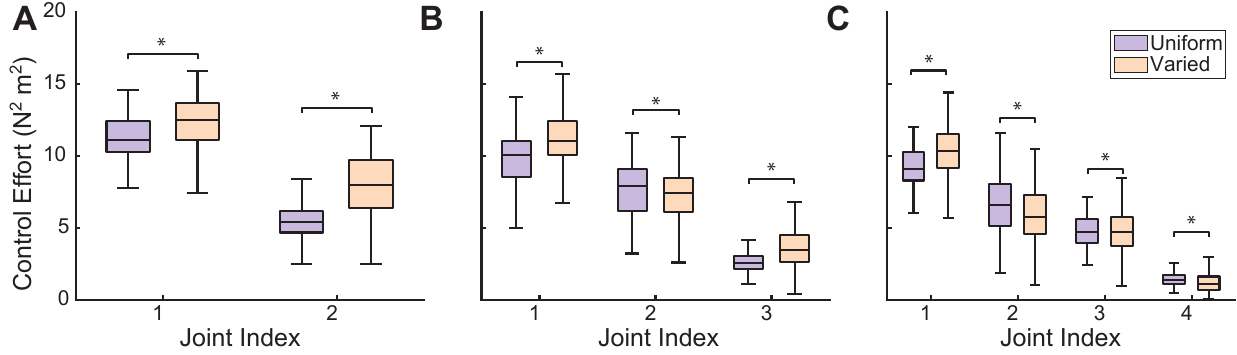}
    \caption{{\bf Comparison of Control Effort per Joint.}
    The box plots display the comparison of the integral of the control effort for each joint within a tail across different trials.
    Note that the total control effort across all degrees of freedom is constrained in each time step, but integral of total control effort through time is unconstrained.
    The control effort is calculated by integrating the sum of squared torque inputs for each joint, each of which has 2DOF, corresponding to pitch and yaw.
    (\textbf{a}), (\textbf{b}), and (\textbf{c}) represent the comparisons for the two-vertebrae, three-vertebrae, and four-vertebrae cases, respectively.
     The `*' symbol indicates significance at $\alpha=0.05$, 
     according to a paired t-test.
    }
    \label{fig:joint_control_effort}
\end{figure}

To mechanistically understand how varying vertebral length along a tail could improve inertial performance, we compared the control effort at each joint between models with uniform vertebral lengths and varied vertebral lengths (Fig. \ref{fig:joint_control_effort}). 
In both uniform and variable vertebral length tails, the highest control effort was applied at the base of the tail, decreasing distally.
We also measured the maximum value of the magnitude of velocity at the most distal tip of each tail (Fig. \ref{fig:box_plot_track_err_varied_len} B).
For tails with two and three vertebrae, the maximum tip velocity was significantly lower for the variable vertebral length tail, compared to the equivalent uniform vertebral length tail.
For tails with four vertebrae, the maximum tip velocity of the variable vertebral length tail was significantly higher than its uniform length counterpart.

\subsection*{Mammal tail vertebral patterns}
The number of tail vertebrae varied from 16 (\emph{Crocidura cyanea}) to 33 (\emph{Peromyscus californicus}	and \emph{Rattus rattus}).
The average number of tail vertebrae for IM species was $28.59\pm2.6$, while the average was $25\pm7.6$ for nonspecialist species. 

In all tails measured for the study, we found a crescendo-decrescendo pattern in the length of the bones (raw vertebral length data for each species included in supplemental material), consistent with a previous description of mouse vertebrae \cite{shinohara1999mouse}.
However, there were pronounced differences in the slope of the crescendo between species that are specialized for inertial maneuvering and those without specific tail functional adaptations.
The maximum difference in normalized length between neighboring pairs of vertebrae was significantly higher in animals specialized for inertial maneuvering (mean $0.91$) versus the nonspecialist group (mean $0.30$) (t-test, $p=6*10^{-4}$, $t = 4.48$, $df = 12.60$).
\section*{Discussion}
Our study finds that adding joints fundamentally enhances tail performance as an inertial appendage, and this is achieved by creating complex curves with multiple inflexion points throughout the length of the tail.
This has the effect of dynamically varying tail moment of inertia and increasing acceleration on the tip of the tail. 
However, it is important to note that incorporating additional articulation into the model would increase the design, modeling, and control complexity of any robotic tail inspired by this simulation.
The observed diminishing returns in error reduction underscore a critical point: increasing the number of links is not invariably a better strategy.
There might exist a balance point between achieving high articulation and maintaining manageable complexity in design and control, which is pivotal in enhancing the model's maneuverability without overly complicating its construction and operation.
We expect animals that use their highly jointed tails as inertial appendages to have evolved strategies that reduce the overall control effort for these tasks, such as muscle synergies \cite{ting2005limited} or branched tendons \cite{fu2024arborsim} that allow one muscle to actuate several joints simultaneously.

We were genuinely surprised to find that the morphologies output by the optimization on within-tail bone lengths generated the crescendo-decrescendo pattern that is exaggerated in mammals associated with inertial maneuvering.
A major benefit of our approach is the ability to compare the trajectories output by the uniform and variable length optimizations to mechanistically understand why an exaggerated crescendo might enhance inertial maneuvering.
We found that the control effort in both the uniform and variable length tails is concentrated at the most proximal joint.
This suggests that greater acceleration can be generated at these sections, facilitating rapid initiation and cessation of tail movements. 
In variable length tails, having two joints close together are functionally equivalent to one joint with a larger range of motion, enabling the distal bones to increase velocity even more.
In the comparison between three-vertebra and four-vertebra configurations, the second link---which represents the longest bone within the tail---required smaller control efforts, meaning less torque is needed in this area to modify the tail's movement.
This potentially indicates that the longer lengths of the mid-section vertebrae function to store more rotational energy.
Based on the pattern we observed, in which the tip velocity contributes proportionally more to angular momentum as the number of tail vertebrae increases, we expect the benefits of a crescendo-decrescendo pattern to be even more pronounced in real mammal tails with 30-40 vertebrae.
This benefit may explain why the number of vertebrae in tails of mammals that perform inertial maneuvering is less variable than in the tails of mammals without obvious tail specializations.
This model-based approach can help provide mechanistic explanations for the correlational patterns found in previous functional morphology studies \cite{schmitt2005role,williams2019increased,mincer2020substrate}.

Future work will also incorporate the non-independence of tail joint actuation.
A major assumption of this model is that each joint can be independently actuated.
However, mammal tails are actuated primarily by extrinsic muscles in the torso, which transmit strains through elongate tendons that cross several joints before inserting on caudal vertebrae \cite{hofmann2021squirrel}.
This muscle-tendon architecture couples the actuation of all joints ``upstream'' from the distal insertion point.
We plan to incorporate muscle-tendon modeling techniques to account for these relationships when we examine models of specific animals \cite{fu2024arborsim}.
However, the computational effort of our optimizations currently prevent us from examining tails with the same number of bones as many of these animals.
Our analysis is subject to the curse of dimensionality, which is particularly acute because the rigid body dynamics of these models are nonlinear.
This results in a non-convex optimization problem, in which the number of degrees of freedom and decision variables exponentially increase the difficulty of solving the optimization.
For our tails, each link adds two degrees of freedom, and the variable tail length analysis adds an additional decision variable.
This makes it challenging to guarantee that the given solution is globally optimal.
We navigated this issue by running our optimization many times with different initial conditions.
As the size of the dataset of potentially optimal solutions increases, we are more likely to approach globally optimal solutions.
Our future work will involve finding new ways to formulate these models and compute the optimization to reduce the computational effort and enable modeling for tails that match or succeed the complexity of real mammal tails \cite{bruder2021advantages}.

This study represents the first quantitative assessment and prediction of maximal inertial appendage performance with jointed tails and in 3D space.
The results demonstrate that the optimization-based approach is an effective way to quantitatively assess the maximal performance of diverse inertial appendages.
A major benefit of this approach is that it uses a Unified Robot Descriptor File (URDF), which is a standard file format in robotics that provides a kinematic and dynamic description as well as a visual representation of a rigid body system.
As we have shown here, this file type can describe both robots and animal structures.
Because our approach is so generalizeable, we hope that researchers apply it to study other types of biological (e.g., wings \cite{boerma2019wings}, arms \cite{yeadon1993biomechanics}, legs \cite{burrows2015mantises}) and robotic (e.g., hybrid tail and reaction wheel \cite{chu2023combining}, linearly actuated center of mass tail \cite{an2020development}) inertial appendages in the future.

\section*{Acknowledgments}
The authors would like to thank Daniel B. Johnson for finding the videos to categorize the tail functions.

We would like to thank the following museum curators for their help with gaining access to the specimens: 
Philip Unitt, (San Diego Natural History Museum);
Mark Omura (Harvard Museum of Comparative Zoology);
Kayce Bell (Natural History Museum of Los Angeles County).

TYM was funded by an Oak Ridge Associated Universities Ralph E. Powe Junior Faculty Enhancement Award.
CJW was funded by a Ruth L. Kirschstein National Research Service Award Individual Postdoctoral Fellowship (5F32AR079923-03).

\nolinenumbers 

\bibliographystyle{IEEEtran}
\bibliography{references}

\end{document}